\documentclass[10pt,twocolumn]{article} 
\usepackage{simpleConference}
\usepackage{times}
\usepackage{graphicx}
\usepackage{amssymb}
\usepackage{url,hyperref}
\usepackage[utf8]{inputenc} 
\usepackage[T1]{fontenc}    
\usepackage{hyperref}       
\usepackage{url}            
\usepackage{booktabs}       
\usepackage{amsfonts}       
\usepackage{nicefrac}       
\usepackage{microtype}      
\usepackage{lipsum}
\usepackage{graphicx}
\usepackage{siunitx}
\graphicspath{ {./images/} }

\usepackage{graphicx}%
\usepackage{multirow}%
\usepackage{amsmath,amssymb,amsfonts}%
\usepackage{amsthm}%
\usepackage{mathrsfs}%
\usepackage[title]{appendix}%
\usepackage{xcolor}%
\usepackage{textcomp}%
\usepackage{manyfoot}%
\usepackage{booktabs}%
\usepackage{algorithm}%
\usepackage{algorithmicx}%
\usepackage{algpseudocode}%
\usepackage{listings}%

\title{Probing the Limits and Capabilities of Diffusion Models for the Anatomic Editing of Digital Twins}

\author{
    \begin{tabular}{cccc}
        Karim Kadry & Shreya Gupta & Farhad R. Nezami & Elazer R. Edelman \\
        MIT & MIT & Brigham and Women's Hospital & MIT \\
        kkadry@mit.edu & shreyag@mit.edu & frikhtegarnezami@bwh.harvard.edu & ere@mit.edu \\
    \end{tabular}
}

\begin{document}
\maketitle

\begin{abstract}
Numerical simulations can model the physical processes that govern cardiovascular device deployment. When such simulations incorporate digital twins; computational models of patient-specific anatomy, they can expedite and de-risk the device design process. Nonetheless, the exclusive use of patient-specific data constrains the anatomic variability which can be precisely or fully explored. In this study, we investigate the capacity of Latent Diffusion Models (LDMs) to edit digital twins to create anatomic variants, which we term digital siblings. Digital twins and their corresponding siblings can serve as the basis for comparative simulations, enabling the study of how subtle anatomic variations impact the simulated deployment of cardiovascular devices, as well as the augmentation of virtual cohorts for device assessment. However, while diffusion models have been characterized in their ability to edit natural images, their capacity to anatomically edit digital twins has yet to be studied. Using a case example centered on 3D digital twins of cardiac anatomy, we implement various methods for generating digital siblings and characterize them through morphological and topological analyses. We specifically edit digital twins to introduce anatomic variation at different spatial scales and within localized regions, demonstrating the existence of bias towards common anatomic features. We further show that such anatomic bias can be leveraged for virtual cohort augmentation through selective editing, partially alleviating issues related to dataset imbalance and lack of diversity. Our experimental framework thus delineates the limits and capabilities of using latent diffusion models in synthesizing anatomic variation for \textit{in silico} trials.
\end{abstract}


\section{Introduction}
\begin{figure}[h!!!]
    \centering
    \includegraphics[width=\columnwidth]{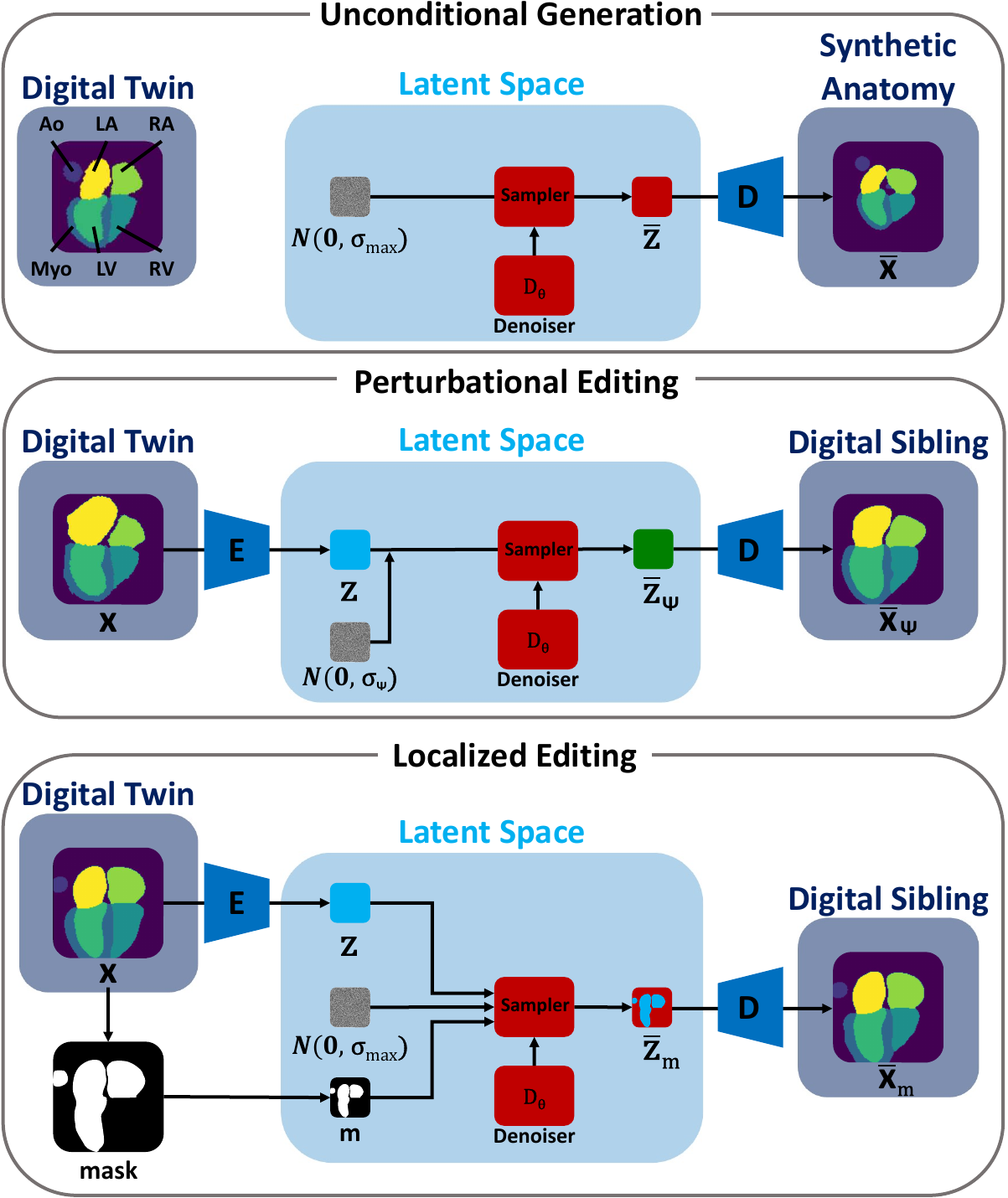}
    \caption{We study the ability of diffusion models to generate digital siblings for virtual interventions and augment in silico trials. Top row: we unconditionally generate latent codes ($\mathbf{\bar{z}})$ which are decoded (\textbf{D}) into cardiac label maps ($\bar{\mathbf{x}}$). Middle row: We encode (\textbf{E}) patient-specific digital twins ($\mathbf{x}$) into a latent space ($\mathbf{z})$ and apply a partial perturb-denoise process to achieve scale-specific variations ($\bar{\mathbf{x}}_\psi$). Bottom row: We locally edit pre-specified tissues to achieve region-specific variations ($\bar{\mathbf{x}}_\mathbf{m}$).}    
  \label{main_fig}
\end{figure}
Physics-based simulations of cardiovascular interventions such as endovascular stent expansion or heart valve implantation can help optimize device design and deployment, especially in challenging anatomies \cite{sarrami2021silico}. These ``virtual interventions'' can be modeled on a patient-specific digital twin, which is a computational replication of a real anatomy derived from medical imaging \cite{kadry2021platform, rouhollahi2023cardiovision,straughan2023fully}. Virtual interventions have been shown to model the mechanical and hemodynamic consequences of implanting heart valves \cite{bianchi2019patienttavr,kusner2021understanding}, atrial appendage occluders \cite{ranard2022feopsaao}, and coronary stents \cite{karanasiou2020designstent,conway2021acute}, as well as the electrophysiological consequences of cardiac ablation \cite{roney2020silicoablation}. Applied to a cohort of digital twins, virtual interventions enable \textit{in silico} trials of medical devices \cite{viceconti2021possibleinsilico}, in which their mechanical safety and efficacy can be assessed within a digital environment. Such trials can accelerate medical device development and de-risk novel designs, potentially reducing the exorbitant cost and failure rates involved with bringing a device to market \cite{sertkaya2022estimatedcost,niederer2020creation}.

Virtual interventions also enable the simulation of hypothetical scenarios, such as implanting alternative devices or modeling different physiological conditions within the same patient \cite{sarrami2021silico}. This experimental framework provides mechanistic insight regarding what factors concerning device design and physiology critically influence deployment. Such insights can enhance the design process and help guide clinical trial recruitment \cite{sarrami2021silico}. In contrast, our ability to simulate insightful scenarios involving alternative anatomic variants is highly limited. Specifically, we delineate three phenomena critical to device development and evaluation that digital twin frameworks are unable to properly address. First, the uniqueness of each digital twin complicates the assessment of uncertainty in device performance attributable to scale-specific anatomic variation. Small scale anatomic features can be highly influential on both hemodynamics and biomechanics. Examples include coronary plaque rupture being influenced by thin fibrous caps \cite{fabris2022thinsmall}, ventricular trabeculae influencing cardiac hemodynamics \cite{sacco2018leftsmall}, and coronary branches affecting blood-flow through the aortic root \cite{moore2015coronarysmall}. Second, due to the complex correlations between local anatomic features within digital twin cohorts, it remains difficult disentangle the causal relationships and interaction effects exerted by localized anatomic regions on device failure. Localized anatomic features have been widely known to interact in influencing cardiovascular physics, examples include the interactions between lipid and calcium in determining plaque rupture risk \cite{kadry2021platform,straughan2023fully}, mitral valve pathology on trans-catheter aortic valve replacements \cite{keshavarz2020mixed_interact}, and trans-catheter aortic valve replacements on coronary flow \cite{tavrgarber2023impact_interact}. Lastly, the reliance on digital twin cohorts for \textit{in silico} trials can compromise device evaluation on less common or pathological anatomic shapes \cite{viceconti2021possibleinsilico,fogel2018factors}. Accordingly, current digital twin paradigms are unable to fully or precisely explore anatomic space, limiting the broader applicability of virtual interventions for device development.

In this study, we investigate the use of latent diffusion models (LDMs) as a controllable source of anatomic variants for \textit{in silico} trials to fulfill two main functionalities. The first functionality centers on the controlled synthesis of informative anatomies through editing digital twins, which we term "digital siblings". As opposed to a digital twin, which is a computational replication of a patient-specific anatomy, a digital sibling would resemble the corresponding twin, but exhibit subtle differences in anatomic form. Comparative simulation studies using twins and their siblings would thus yield insight regarding how scale-specific and region-specific anatomic variation can influence simulated deployment. The second functionality revolves around virtual cohort augmentation by creating digital siblings from a curated subpopulation of digital twins. This would enrich virtual cohorts with specified anatomic attributes, addressing issues related to cohort imbalance and diversity. We accordingly develop a latent diffusion model to generate 3D cardiac label maps and introduce a novel experimental framework to study the synthesis of anatomic variation (Fig \ref{main_fig}). We first characterize the baseline performance of the model through generating de-novo cardiac label maps. We then investigate two methods to generate digital siblings with diffusion models: 1) perturbational editing of cardiac digital twins to enable scale-specific variation; and 2) localized editing of cardiac digital twins to enable region-specific variation. In our experimental framework, we select various digital twins to act as "seed" volumes and produce several digital siblings through editing. We then apply this procedure over different hyperparameters and seed characteristics to study how generative editing can alter the morphological and topological attributes of digital twins. Lastly, we study how such editing methods can be used to augment virtual cohorts with less common anatomic features. Our main contributions and insights are as follows:
\begin{enumerate}
    \item We develop and train a latent diffusion model to generate 3D cardiac label maps and  introduce a novel experimental framework to study how generative editing techniques can produce scale-and-region specific variants of digital twins.
    \item We demonstrate that latent diffusion models can introduce topological violations during generation and editing, where the number of violations is influenced by editing methodology and seed characteristics.
    \item We find that dataset imbalance induces a bias within the generation process towards common anatomic features. This anatomic bias extends to scale-and-region specific editing. The degree and spatial distribution of this bias is influenced by editing hyperparameters and seed characteristics.
    \item We demonstrate that this anatomic bias can be leveraged to enhance virtual cohort diversity in two manners. Virtual cohort augmentation with scale-specific variation can help explore less populated spaces within the anatomic distribution bounded by the training set, while augmentation with region-specific variation can augment the cohort with anatomic forms outside the anatomic distribution.
\end{enumerate}

\section{Related Work}
\begin{figure*}[h!!!]
    \centering
    \includegraphics[width=\textwidth]{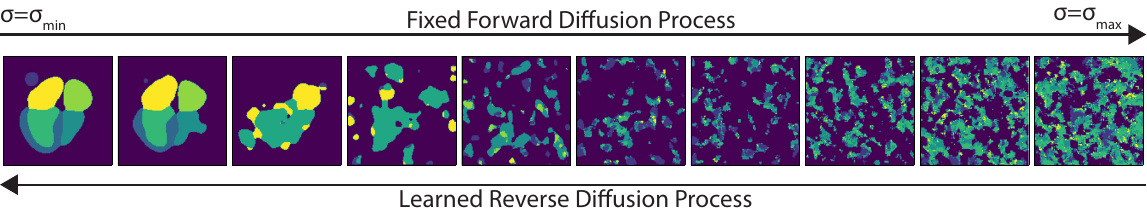}
    \caption{Schematic for the forward and reverse diffusion process showing the decoded cardiac label maps for several intermediately noised latent representations $\mathbf{z}_\sigma$. During training, a neural denoiser learns to approximate the reverse process at each noise level $\sigma$. During sampling, the network is recursively applied  to produce de-novo cardiac label maps.}    
  \label{Diffusion_Fig}
\end{figure*}
\subsection{Generative models of Virtual Anatomies}
Generative models of virtual anatomies typically struggle to balance between producing outputs that are realistic with those that can be controlled by the user. The gold standard method is Principal Components Analysis (PCA), which has traditionally been used to generate virtual cohorts for biomechanical and hemodynamic simulations \cite{williams2022aortic}. Despite its utility, PCA is unable to accurately model the highly nonlinear anatomic variation inherent to human anatomy. As such, there has been a rising interest in deep learning approaches for producing virtual anatomies. State-of-the-art deep learning architectures for this purpose have been variational autoencoders (VAEs) and generative adversarial networks (GANs), which exhibit improved performance if trained with a sufficiently large dataset \cite{beetz2022interpretablecardiacvae,beetz2021generatingconditional,qiao2023cheart}. While such architectures have demonstrated the ability to produce variations of anatomy by exploring their latent space \cite{beetz2022interpretablecardiacvae}, as of yet current approaches are limited in their ability to precisely edit patient-specific models. As such, previous approaches cannot controllably introduce anatomic variation at different spatial scales or within localized regions while keeping others constant.

\subsection{Diffusion Models for Human Anatomy}
Diffusion models have been shown to produce 2D and 3D medical images with high quality \cite{pinaya2022brain,muller2022diffusionbeatsganmed,khader2023denoising}. However, the use of diffusion models to generate virtual anatomies in the form of anatomic label maps is still in its infancy, with most studies focusing on generating label maps for the purpose of training downstream computer vision algorithms. Preliminary studies utilized  unconditional diffusion models to produce 2D multi-label segmentations of both the brain and retinal fundus vasculature respectively \cite{fernandez2022can,go2023generationretinalfundus}. However, they did not directly evaluate the generated virtual anatomies with respect to morphological or topological quality, factors that are critical to their use within \textit{in silico} trials.

\subsection{Diffusion Models for Generative Editing}
The ability of diffusion models to flexibly edit natural images is well-characterized. For example, diffusion models can create variations of natural images through a perturb-denoise process, partially corrupting a seed image and restoring it through iterative denoising \cite{ho2020denoising}. The level of added noise can control whether the model synthesizes global or local features \cite{meng2021sdedit}. Additionally, diffusion models can be used to locally in-paint regions within an image by specifying a spatially extended mask \cite{nichol2021glide}. This technique has been used in the context of medical images for anomaly detection \cite{bercea2023maskanomaly,fontanella2023diffusionanomaly} and data augmentation for brain images \cite{rouzrokh2022multitaskinpaint}. However, such techniques have not been characterized in the context of generating anatomic variation for virtual interventions, making it difficult to gauge the extent to which locally editing or restoring a partially corrupted anatomy introduces morphological bias or topological defects. Furthermore, in the context of augmenting \textit{in silico} trials, the benefits of  editing over unconditional sampling has not been investigated.

\subsection{Evaluation of Generated Anatomy}
Current methods for generative models are not suited to evaluate the quality of synthetic cohorts for \textit{in silico} trials. For example, the Fréchet Inception Distance (FID) \cite{ho2022classifier} is difficult to use for evaluating generative models of virtual anatomies, as no standard pre-trained network for 3D anatomic segmentations is available. Similarly, pointcloud-based metrics such as minimum matching
distance and coverage are used to evaluate the realism and diversity of generated shapes that exhibit the same topological structure \cite{achlioptas2018learningcovmmd}, but cannot be used on multi-component anatomy with varying topology. Moreover, previously mentioned metrics do not measure interpretable morphological metrics necessary to understand device performance, nor would it measure topological correctness, a critical factor to ensure compatibility with numerical simulation. Recent studies evaluated the plausibility of virtual anatomies by visualizing the 1D distributions of clinically relevant variables such as tissue volumes \cite{romero2021clinicallymetrics,qiao2023cheart}, but fail to study the multi-dimensional relationship between morphological metrics, nor do they investigate morphological bias due to imbalanced data distributions.

\section{Methods}

\subsection{Dataset}
\label{dataset}
We used the TotalSegmentator dataset \cite{wasserthal2022totalsegmentator}, consisting of 1204 Computed Tomography (CT) images, each segmented into 104 bodily tissues. We filtered out all patient label maps that do not have complete and adequate-quality segmentations for all four cardiac chambers. This resulted in a dataset of 512 3D cardiac label maps, where each label map consisted of 6 tissues: aorta (Ao), myocardium (Myo), right ventricle (RV), left ventricle (LV), right atrium (RA), and left atrium (LA). All cardiac label maps were cropped and resampled to a size of $7 \times 128 \times 128 \times 128$, with an isotropic voxel size of \SI{1.4}{\cubic\mm}. We then reoriented each cardiac segmentation so that the axis between the LV and LA centroids is aligned with the positive z-axis. Lastly, we rigidly registered all segmentations to a reference label map using ANTS \cite{avants2008symmetric}.

\subsection{Latent Diffusion Models}
\label{ldm_arch}
We employed a latent diffusion model (LDM), consisting of a variational autoencoder (VAE) and a denoising diffusion model. The VAE encodes cardiac label maps $\mathbf{x}$ into  latent representations $\mathbf{z}$, which can be decoded into label maps $\bar{\mathbf{x}}$. The training process for our diffusion model is done in the latent space of the trained autoencoder, we represent the probability distribution of cardiac anatomy by $p_{data}(\mathbf{z})$ and consider the joint distribution $p(\mathbf{z};\sigma)$ obtained through a forward diffusion process, in which i.i.d Gaussian noise of standard deviation $\sigma$ is added to the data, where at $\sigma=\sigma_{max}$ the data is indistinguishable from Gaussian noise. The driving principle of diffusion models is to sample pure Gaussian noise and approximate the reverse diffusion process through using a neural network to sequentially denoise the latent representations $\mathbf{z}_\sigma$ with noise levels $\sigma_0=\sigma_{max}>\sigma_1>\dots>\sigma_N=\sigma_{min}$ such that the final denoised latents correspond to the clean data distribution. Following Karras et al. \cite{karras2022elucidating}, we represent the reverse diffusion process as the solution to the following ordinary differential equation

\begin{equation}
d\mathbf{z} = -  \sigma \nabla_\mathbf{z} \log p(\mathbf{z}; \sigma) \, dt
\label{eq:ode}
\end{equation}
Where the score function $\nabla_\mathbf{z} \log p(\mathbf{z;\sigma})$ denotes the direction in which the rate of change for the log probability density function is greatest. Since the data distribution is not analytically tractable we train a neural network to approximate the score function. We start with clean latent representations $\mathbf{z}$ and model a forward diffusion process that produces intermediately noised latents $\mathbf{z}_\sigma=\mathbf{z}+\mathbf{n}$ where $\mathbf{n} \sim \mathcal{N}(\mathbf{0},\sigma^2\mathbf{I})$, parameterized by a noise level $\sigma$. The diffusion model is parameterized as a function $F_{\theta}$, encapsulated within a denoiser $D_{\theta}$, that takes as input an intermediately noised output $\mathbf{z}_\sigma$ and a noise level $\sigma$ to predict the clean data $\mathbf{z}$.

\begin{equation}
    D_{\theta}(\mathbf{z_\sigma};\sigma) = 
    c_{\text{skip}}(\sigma)\,\mathbf{z_\sigma} + 
    c_{\text{out}}(\sigma)\,F_{\theta}(c_{\text{in}}(\sigma) \, \mathbf{z_\sigma}; \, c_{\text{noise}}(\sigma)) \,,
\end{equation}

where $c_{\text{skip}}$ controls the skip connections that allow the $F_{\theta}$ to predict the noise $\mathbf{n}$ at low $\sigma$ and the training data $\mathbf{z}$ at high $\sigma$. The variables $c_{\text{out}}$ and $c_{\text{in}}$ scale the input and output magnitudes to be within unit variance, and the constant $c_{\text{noise}}$ maps the noise level $\sigma$ to a conditioning input to the network \cite{karras2022elucidating}. The denoiser output is related to the score function through the relation $\nabla_\mathbf{z} \log p(\mathbf{z}; \sigma) = \left( D_\theta(\mathbf{z_\sigma}; \sigma) - \mathbf{z} \right) / \sigma^2$ and $F_{\theta}$ is chosen to be a 3D U-net with both convolutional and self-attention layers, similar to previous approaches \cite{fernandez2022can,karras2022elucidating,ho2020denoising,rombach2022ldm}. Full details on the VAE and U-net architectures can be found in appendix \ref{appdx:diffusion_model}. The loss $L$ is then specified based on the agreement between the denoiser output and the original training data:
\begin{equation}
    L=\mathbb E_{\sigma,\mathbf{z},\mathbf{n}}[\,\lambda(\sigma)\vert\vert D_\theta(\mathbf{z}_\sigma;\sigma)\,-\mathbf{z}\vert\vert^2_2] \,,
\end{equation}

such that the loss weighting $\lambda(\sigma)=1/c_{\text{out}}(\sigma)^2$ ensures an effective loss weight that is uniform across all noise levels, and $\sigma$ is sampled from a log-normal distribution with a mean of 1 and standard deviation of 1.2. 

Once the denoiser has been sufficiently trained, we define a specific noise level schedule governing the reverse process, in which the initial noise level, $\sigma$, starts at $\sigma_{\text{max}}$ and decreases to $\sigma_{\text{min}}$:
\begin{equation}
    \sigma_{i}=\left(\sigma_{\text{max}}^\frac{1}{\rho}+\frac{i}{N-1}(\sigma_{\text{min}}^\frac{1}{\rho}-\sigma_{\text{max}}^\frac{1}{\rho})\right)^\rho 
\end{equation}
where $\rho, \sigma_{min}$ and $\sigma_{max}$ are hyperparameters that were set to 3, 2e-3, and 80 respectively. We specifically leverage the deterministic sampling algorithm detailed in Karras et al. \cite{karras2022elucidating} to sequentially denoise the latent representations $\mathbf{z}_\sigma$ and solve the reverse diffusion process detailed in Eq. \ref{eq:ode} (Figure \ref{main_fig}).

\begin{figure*}[h!!!]
\centering\includegraphics[width=\linewidth]{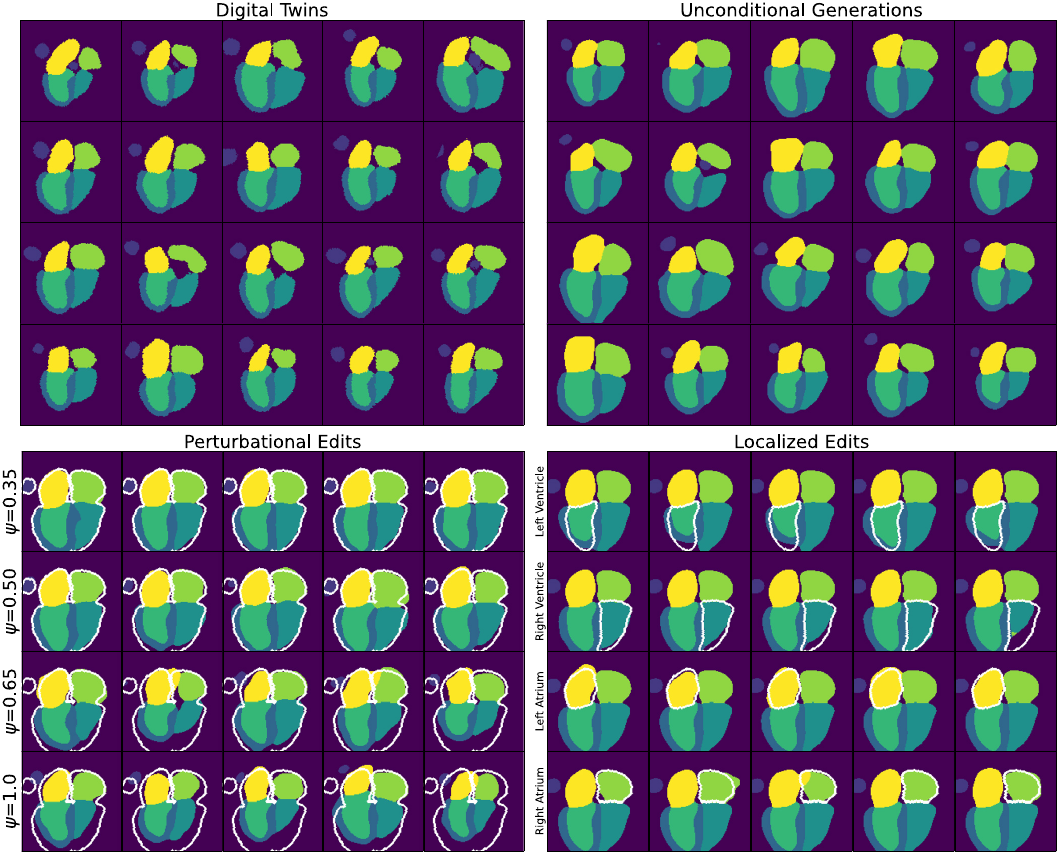}
    \caption{Example 2D slices from 3D cardiac label maps. Top left: digital twin label maps from the training set. Top right: unconditionally generated label maps generated by the diffusion model. Bottom left: perturbational edits of a single cardiac digital twin over various sampling ratios. Bottom right: localized edits of cardiac digital twins over various tissue masks. Bottom row has a white outline of the edited twin for perturbational edits (left) and an outline of the edited tissue region for localized edits (right).}
  \label{uncond_cardiac_grid}
\end{figure*}

\begin{figure*}[h!!!]
  \centering
  \begin{minipage}{0.37\textwidth}
    \includegraphics[width=\textwidth]{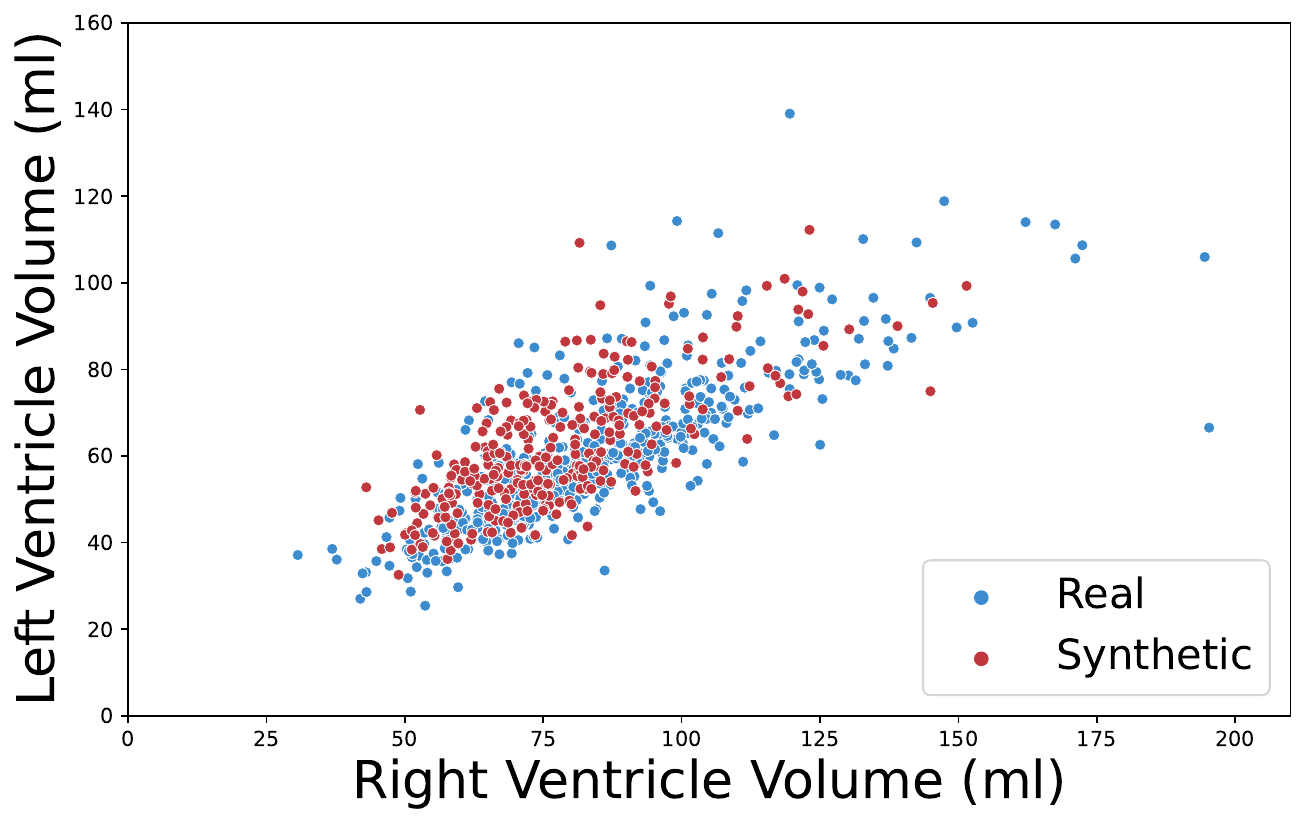}
    \caption{Unconditional generation captures common anatomic variations but fail to capture outliers. Scatterplot shows the 2D morphological distribution exhibited by real cohorts and synthetic cohorts generated by unconditional sampling.}

    \label{uncond_pairplot}
    
  \end{minipage}
  \hfill 
  \begin{minipage}{0.55\textwidth}\
  \includegraphics[width=\textwidth]{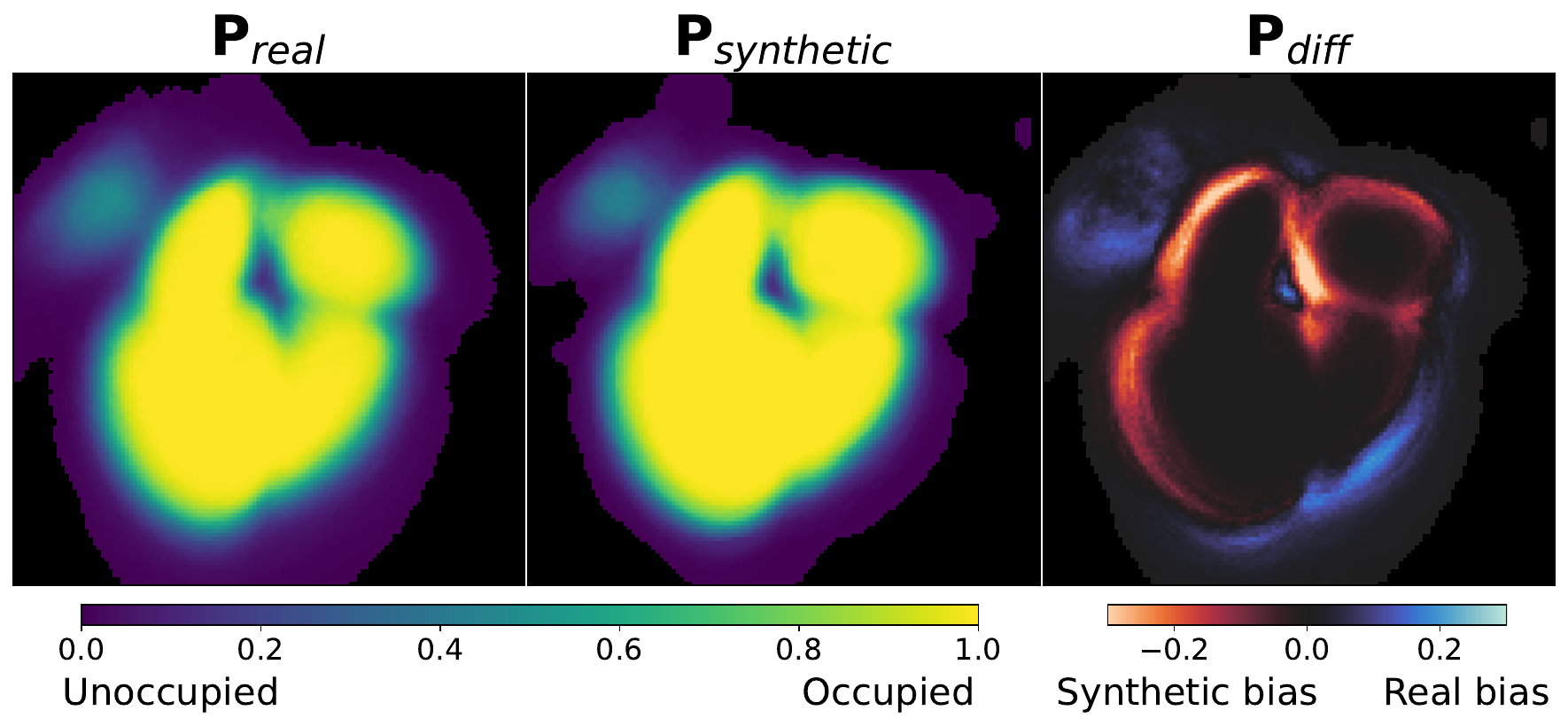}
    \caption{The distribution of synthetic label maps exhibits spatially variant discrepancies against that of real label maps. Spatial occupancy heatmaps show the distribution of real ($\mathbf{P}_{\text{real}}$) and synthetic ($\mathbf{P}_{\text{synthetic}}$) label maps, as well as the difference in occupation ($\mathbf{P}_{\text{diff}}$) . Heatmaps are masked out where $\mathbf{P}_{\text{real}}$ or $\mathbf{P}_{\text{synthetic}}$ are zero. Real or synthetic bias correspond to increased relative occupancy by real or synthetic anatomies respectively.}
    
    \label{uncond_hm}
  \end{minipage}
  \label{fig:uncond_combined}
\end{figure*}

\subsection{Perturbational Editing}
\label{perturbation}
To create digital siblings by perturbational editing, we first encoded a seed cardiac label map $\mathbf{x}$ into the latent representation $\mathbf{z}$. Instead of sampling from pure Gaussian noise, we  recursively apply the denoiser using the intermediately noised latent $\mathbf{z}_\sigma$ as the starting point (Figure \ref{main_fig}) to produce $\bar{\mathbf{z}}_\psi$. The latent $\bar{\mathbf{z}}_\psi$ is then decoded into the cardiac label map $\bar{\mathbf{x}}_\psi$ using the autoencoder. The intermediate step $i<N$ is a hyperparameter that determines how much of the sampling process is recomputed and is parametrized as the sampling ratio $\psi=(i-N)/N$ in our experiments.

\subsection{Localized Editing}
\label{editing}
To create digital siblings by localized editing, we first encoded a seed cardiac label map $\mathbf{x}$ into the latent representation $\mathbf{z}$. A tissue-based mask, $\mathbf{m}$, denoting which cardiac tissues are to be preserved, was created and downsampled to the same size as the latent representation. The mask was then dilated twice to ensure that tissue interfaces remain stable during editing. The sampling process is similar to that of unconditional sampling, with the addition of an update step that replaces the unmasked portion of the intermediately denoised image with an equivalently corrupted latent representation belonging to the seed label map:
\begin{equation}
    \mathbf{z}_\sigma= (\mathbf{z}+ \mathbf n(\sigma))*\mathbf{m} + (1 - \mathbf{m}) * \mathbf{z}_\sigma \,.
\end{equation}
At the end of sampling, the denoised latent $\bar{\mathbf{z}}_\mathbf{m}$ is then decoded into the cardiac label map  $\bar{\mathbf{x}}_\mathbf{m}$ through the decoder (Figure \ref{main_fig}).

\subsection{Evaluating Morphology and Topology}
To assess the morphological quality of a virtual cohort, we represented each virtual anatomy in terms of a 12-dimensional morphological feature vector. For each cardiac label map, we calculate the volume, major axis length, and minor axis length for the LV, RV, LA, and RA. Two of these metrics (LV and RV volumes) were further chosen to plot the global morphological distribution of each cohort. To quantitatively evaluate the morphological similarity of two virtual cohorts, we calculated improved precision and recall \cite{kynkaanniemi2019improved}, as well as Fréchet distance using morphological vectors that were normalized by the mean and standard deviation of the real dataset values. Precision and Fréchet distance would measure the anatomic fidelity of the generated anatomies, while recall would measure their diversity. To visualize the anatomic bias on a local scale, a voxel-wise mean was computed over all virtual anatomies within a cohort. This results in a spatial heatmap $\mathbf{P}$ of size $7 \times 128 \times 128 \times 128$ for the real and synthetic cohorts. The inverse of the background channel was chosen for further visualization.

Furthermore, in order to study how well anatomic constraints and compatibility with numerical simulation are respected, we assess the topological quality of each label map. Clinically, topological defects such as a septal defect between the right and left hearts can have a significant effect on electrophysiology \cite{williams2018arrhythmiastopo} and hemodynamics \cite{shah2017impactatrialtopo}. Specifically, for each generated anatomy we evaluate 12 different topological violations and calculate the percentage of topological violations exhibited by the cohort. Full details on topological assessment can be found in appendix \ref{appdx: topo_metrics}.

\section{Experiments}

\subsection{Unconditional Sampling of Virtual Anatomies}
We conducted a sensitivity analysis of cohort quality with respect to the sampling steps and cohort size and found diminishing returns after 20 sampling steps and a cohort size of 50 (details in appendix \ref{appdx: Sens}). We then sample 360 label maps with 20 steps for analysis and visualization.  Example label maps can be seen in Figure \ref{uncond_cardiac_grid}. The scatterplot (Figure \ref{uncond_pairplot}) and the difference heatmap (Figure \ref{uncond_hm}) show the morphological distribution of the synthetic anatomies on a global and local scale respectively. Both figures demonstrate that unconditional sampling tends to generate mean-sized cardiac label maps, but fails to sample rarer anatomic configurations on the periphery of the distribution. This bias also exists on a local level as seen in the difference heatmap $\mathbf{P}_{\text{diff}}$ in Figure \ref{uncond_hm}.

Table \ref{table_uncond_topo} indicates the primary source of topological violation stems from the initial segmentations and the sampling process, rather than the autoencoder. Violations in the real dataset stem from the segmentation network used to create the original dataset, in which small clusters of misclassified tissues contribute to the amount of topological violations (Figure \ref{uncond_cardiac_grid}). The autoencoded data has a reduced number of topological violations due to inability to reconstruct such small clusters. The diffusion-related violations, in contrast, arise from the inability to preserve fine details during generation, resulting in topological violations such as atrial contact or LV tissue at the RV apex (Figure \ref{uncond_cardiac_grid}).

\begin{figure}[t!!!]
    \centering\includegraphics[width=\columnwidth]{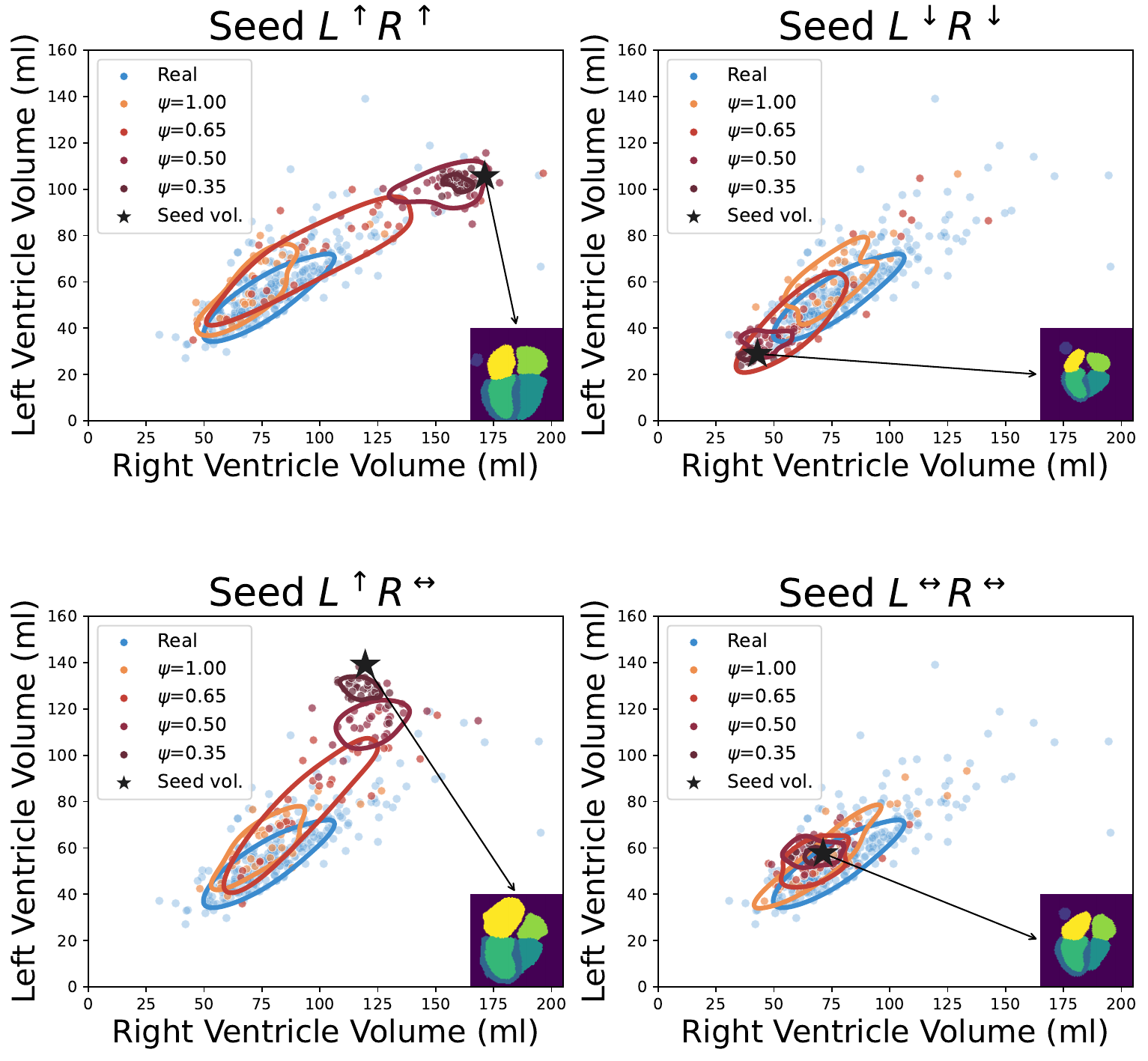}
    \caption{Perturbationally editing seed cardiac label maps (star marker) with increasing levels of injected noise  $\psi$ produces cohorts that are biased towards the most common anatomies (blue contour). Each scatterplot corresponds to a different seed volume, showing multiple cohorts synthesized by editing the same seed with different sampling ratios ($\psi$). For improved visual clarity, scatterplots are supplemented with kernel density estimate plots, and the number of data points displayed per cohort is reduced by half.}

  \label{pert_pairplot}
\end{figure}
\begin{figure}[t!!]
  \centering
    \includegraphics[width=\linewidth]{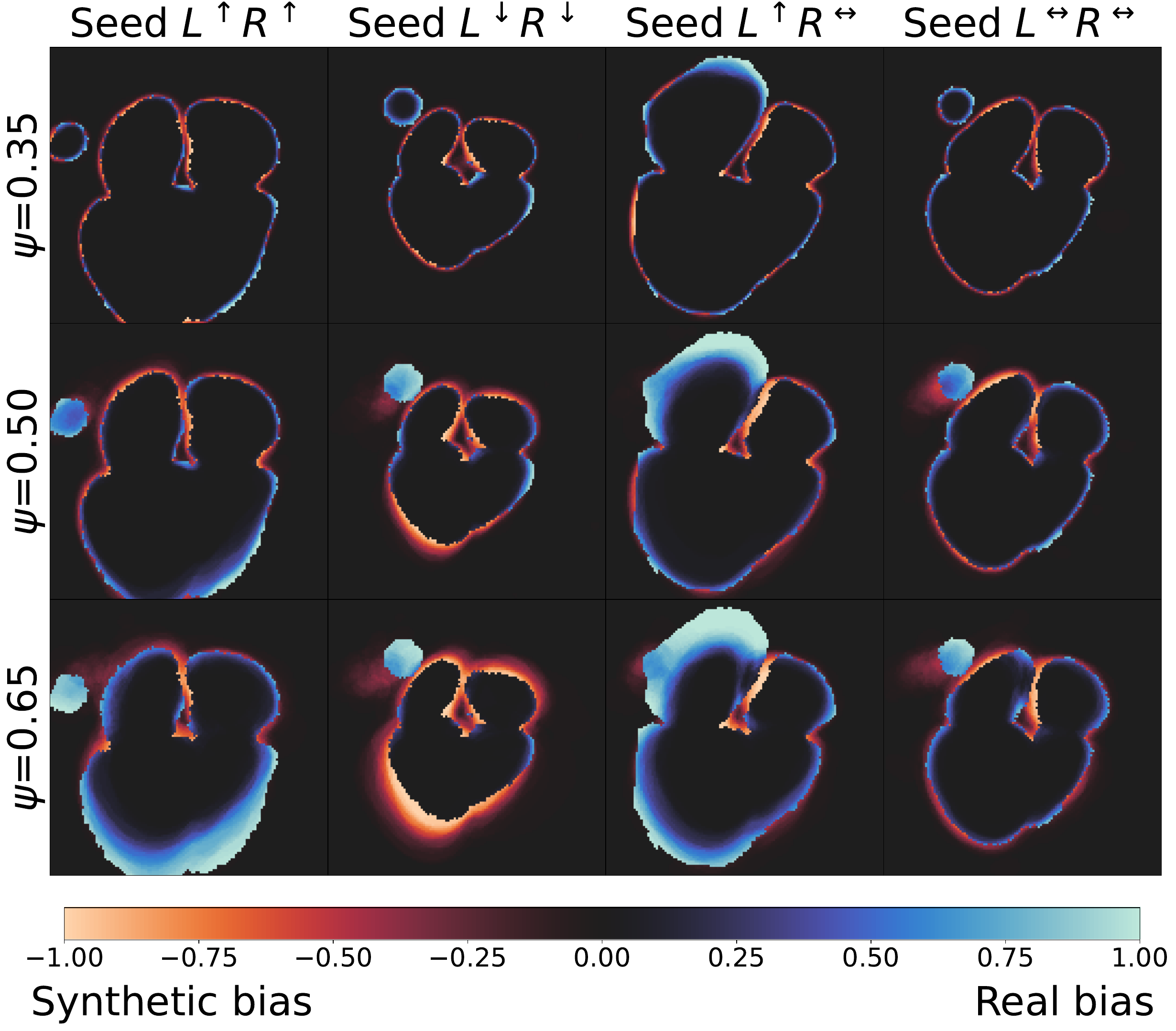}
    \caption{Perturbationally editing seed cardiac label maps (columns) with increasing levels of injected noise $\psi$ (rows) enables scale-specific variation. Difference heatmaps $\mathbf{P}_{\text{diff}}$ show spatially varying discrepancies between the seed and synthetic cohorts generated by perturbationally editing various seed label maps.}
  \label{pert_heatmaps}
\end{figure}

\begin{table}

\centering
\begin{tabular}{lccc}
\hline
 & \text{Real} & \text{Real-VAE} & \text{Synthetic}  \\
\hline
\text{TV (\%)} & 21.0 & 13.7 & 18.0  \\
\hline

\end{tabular}
\caption{Topological violations exhibited by real, autoencoded real, and synthetic cohorts respectively.}
\label{table_uncond_topo}
\end{table}

\subsection{Scale Specific Variation Through Perturbational Editing}
We select four seed label maps that represent different types of cardiac anatomy: a seed with a large LV and RV ($L^\uparrow$ $R^\uparrow$), a seed with a small LV and RV ($L^\downarrow$ $R^\downarrow$), a seed with a large LV but mean sized RV ($L^\uparrow R^\leftrightarrow$), and a seed with a mean sized LV and RV ($L^\leftrightarrow$ $R^\leftrightarrow$). For each seed, we generate synthetic anatomies with varying sampling ratios, corresponding to $\psi$=[0.35, 0.50, 0.65, 0.8, 1], leading to a total of 20 virtual cohorts of 60 anatomies each. Example label maps can be seen in Figure \ref{uncond_cardiac_grid}.

Figure \ref{pert_pairplot} shows that the cohorts generated by perturbational editing are increasingly biased towards the most common anatomies with increasing noise. Figure \ref{pert_heatmaps} further shows that the amount of injected noise corresponds to spatial scale, as the bias exhibited by the spatial heatmap $\mathbf{P}_{\text{diff}}$ expands with increasing noise. Table \ref{table_pert_topo} demonstrates that the topological quality of the generated heatmaps gradually conforms to the synthetic data average after $\psi=1.0$. However, topological quality can degrade when editing outlier twins, as can be seen when perturbationally editing seed $L^\uparrow R^\leftrightarrow$, which occupies a sparsely populated region of the anatomic distribution.

\begin{figure}[h] 
  \centering\includegraphics[width=\columnwidth]{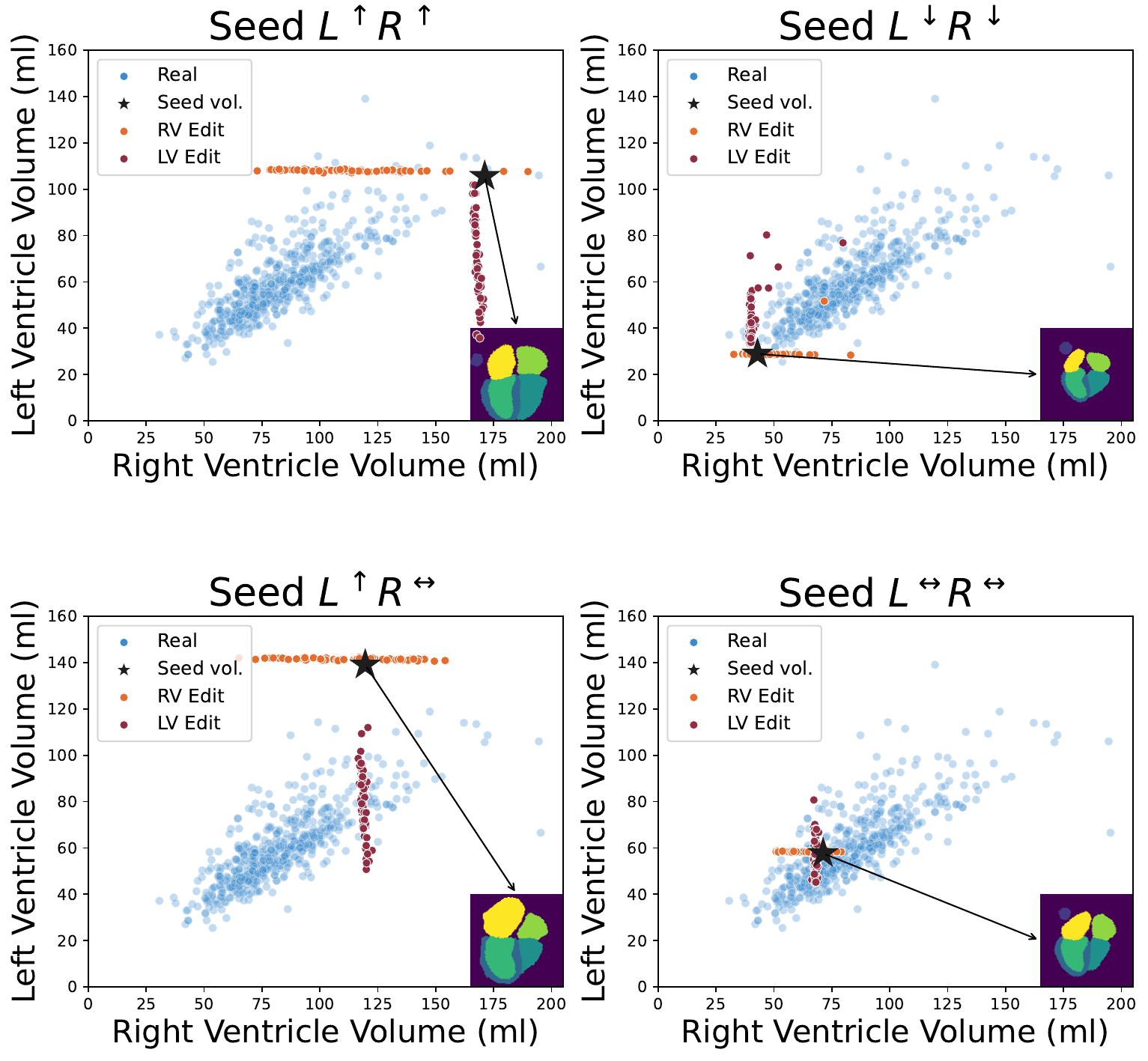}
    \caption{Localized editing of seed cardiac label maps (star marker) produces cohorts with region-specific variation that is biased towards those of the most common anatomies. Each scatterplot corresponds to a different seed, showing multiple cohorts synthesized by locally editing the same seed volume with different tissue masks \textbf{m}.}

  \label{edit_pairplot}
\end{figure}

\begin{figure}[h]
      \centering
      \includegraphics[width=\linewidth]{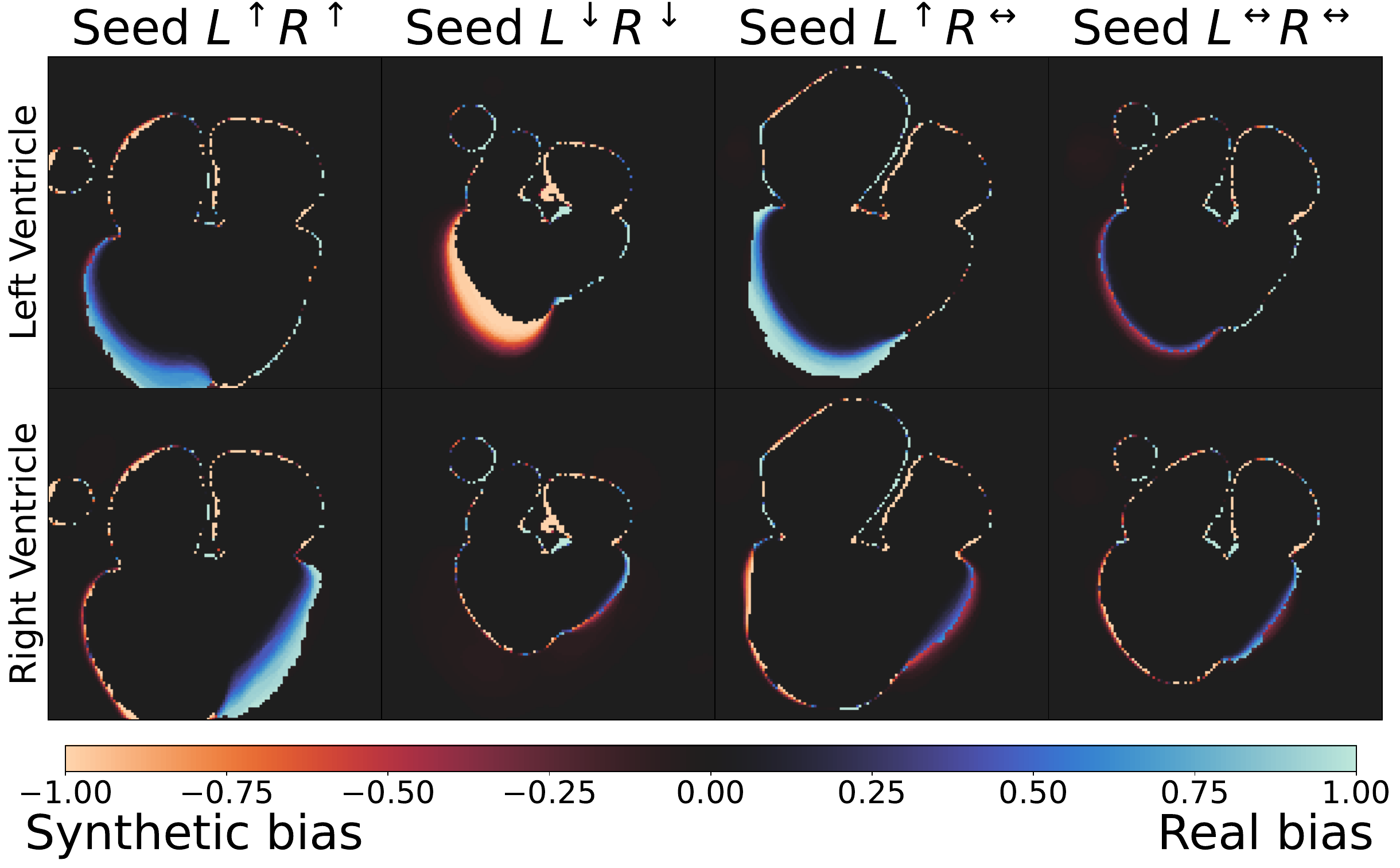}
    \caption{Locally editing seed cardiac label maps (columns) with different tissue masks  $\mathbf{m}$ (rows) enables region-specific variation. Difference heatmaps $\mathbf{P}_{\text{diff}}$ show spatially varying discrepancies between the seed and synthetic cohorts generated by locally editing 4 seed volumes.}
  \label{edit_hm}
\end{figure}

\begin{table}

  \centering
  \begin{tabular}{lcccc}
    \hline
    TV (\%) & $L^\uparrow R^\uparrow$ & $L^\downarrow R^\downarrow$ & $L^\uparrow R^\leftrightarrow$ & $L^\leftrightarrow R^\leftrightarrow$ \\
    \hline
    \text{Original-VAE} & 8.3 & 8.3 & 8.3 & 8.3 \\
    \hline
    \(\psi=0.35\) & 15.1 & 13.8 & 31.8 & 18.5 \\
    \(\psi=0.50\) & 18.3 & 20.3 & 29.6 & 22.2 \\
    \(\psi=0.65\) & 23.6 & 21.3 & 24.9 & 21.9 \\
    \(\psi=0.80\) & 20.6 & 22.9 & 22.1 & 22.6 \\
    \(\psi=1.00\) & 16.4 & 18.8 & 17.4 & 18.3 \\
    \hline
  \end{tabular}
    \caption{Topological violations exhibited by each cohort produced by perturbationally editing various seed label maps for different sampling ratios \(\psi\).}
    \label{table_pert_topo}
\end{table}

\subsection{Region Specific Variation Through Localized Editing}
For each of the previously mentioned seeds, we specify two masks designed to edit the RV and LV respectively. The myocardium was not included for each tissue mask, allowing it to vary with each ventricular chamber. This process resulted in eight synthetic cohorts of 60 anatomies each. Example label maps can be seen in Figure \ref{uncond_cardiac_grid}. 

Figure \ref{edit_pairplot} shows that the 1D distributions of edited ventricular volumes are biased towards most common values of the real cohort. This can be seen most prominently with seed $L^\uparrow R^\leftrightarrow$ where the edited LVs have a substantially lower volume as compared to the seed label map. From the spatial difference heatmaps $\mathbf{P}_{\text{diff}}$ visualized in Figure \ref{edit_hm}, we further observe that localized editing can change individual chambers while maintaining others as constant, where the edited chambers are biased towards a mean anatomic shape. With the exception of editing the RV of seed  $L^\uparrow R^\leftrightarrow$, locally editing the seed label maps did not produce an increased percentage of topological violations as compared to the seeds, as can be seen in Table \ref{table_edit_topo}.

\begin{figure*}[t] 
  \centering\includegraphics[width=\textwidth]{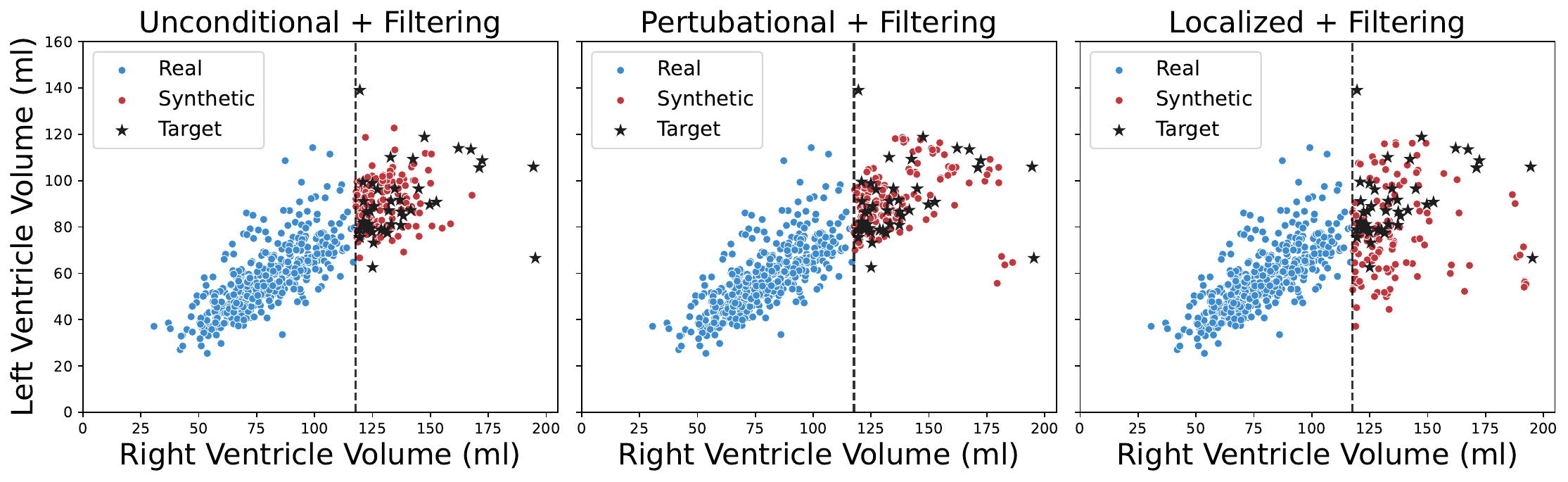}
    \caption{Scatterplots demonstrating three augmentation strategies for a target cohort of real cardiac label maps distinguished by right ventricle volumes larger than a minimum threshold (dashed lines). The first strategy uses unconditional generation while second and third strategies utilized generative editing applied to a cohort of seed label maps. All generated cohorts underwent filtering to ensure a minimum right ventricular volume.}
  \label{cohort_pp}
\end{figure*}

\begin{table}
\centering
\begin{tabular}{lcccc}
\hline
TV (\%) & $L^\uparrow R^\uparrow$ & $L^\downarrow R^\downarrow$ & $L^\uparrow R^\leftrightarrow$ & $L^\leftrightarrow R^\leftrightarrow$ \\
\hline
\text{Original-VAE} & 8.3 & 8.3 & 8.3 & 8.3 \\
\hline
\text{RV Edit} & 12.5 & 10.3 & 16.9 & 10.4 \\
\text{LV Edit} & 10.0 & 11.5 & 9.4 & 9.3 \\
\hline
\end{tabular}
\caption{Topological violations exhibited by each cohort produced by localized editing of various seed label maps with different tissue masks.}
\label{table_edit_topo}
\end{table}

\subsection{Virtual Cohort Augmentation Through Selective Editing}

We contrast and compare three strategies that can augment virtual cohorts with rare anatomies to improve dataset imbalance and diversity. In this case, we enrich a target cohort of rare patient-specific cardiac label maps distinguished by an RV volume larger than a threshold value of 115 ml. Our first strategy is to unconditionally sample 3600 label maps and filter all outputs with RV volumes less the threshold. In our second strategy, we utilize the bias inherent to perturbational editing and modify digital twins from the target cohort to create digital sibling cohorts. Half of the digital twins  received a large perturbation ($\psi$=0.5) and the other half received a small perturbation ($\psi$=0.35). Following the editing process, digital siblings with an RV volume below the threshold were excluded. Our third strategy leverages the bias inherent to localized editing, in which half of the target cohort was locally edited to have different LV shapes, while the other half were edited to have different RV shapes. Similarly, outputs that do not meet the RV volume threshold were excluded.  All three strategies resulted in filtered cohorts of size 140 each. The evaluation metrics, namely Fréchet distance, precision, and recall, were computed against the target cohort consisting of 47 cardiac label maps as the reference standard.

Figure \ref{cohort_pp} demonstrates that unconditional generation does not fully explore the peripheries of the target cohort distribution, where it can be seen that the largest RV volumes within the target cohort are not represented. In contrast, perturbational editing excels in filling sparsely populated peripheries of the distribution. Table \ref{table_cohort_topo} reinforces these insights, demonstrating that augmentation through perturbational editing enhances coverage through exhibiting higher recall values as compared to unconditional generation. Furthermore, both methods yield comparable Fréchet distance and precision values, suggesting equivalent levels of morphological realism. Augmenting cohorts with localized editing yields cardiac label maps with morphological features that conform to the distribution of individual morphological metrics but deviate from the multidimensional distribution. Table \ref{table_cohort_topo} supports this finding, indicating the lowest precision and Fréchet Distance but the highest recall when compared to previous strategies. Table \ref{table_cohort_topo} also demonstrates that virtual cohorts produced by the various augmentation strategies exhibit similar or even enhanced topological quality as compared to filtering the unconditionally generated label maps.

\begin{table}[h]
\centering

  \setlength{\tabcolsep}{2.5pt}
  \begin{tabular}{lcccc}
    \toprule
    \bfseries Augment. Strategy & \bfseries FD & \bfseries Prec. & \bfseries Rec. & \bfseries TV (\%) \\
    \hline
    Uncond. + Filtering & 2.86 & 0.94 & 0.76 & 19.6 \\
    Pert. + Filtering &2.72 & \textbf{0.96}  & 0.93 & 21.6 \\
    Local. + Filtering& \textbf{2.64} & 0.91 & \textbf{0.95} & \textbf{15.1} \\
    \bottomrule

  \end{tabular}
    \caption{Comparison of various metrics across different virtual cohort augmentation strategies. The Fréchet distance, precision, and recall values were calculated using the target cohort as a reference.}
    \label{table_cohort_topo}
\end{table}

\section{Discussion and Conclusions}
In this study we developed an experimental framework to investigate how generative diffusion models can modify anatomic digital twins for virtual interventions. Specifically, we trained a diffusion model on a dataset of 3D cardiac label maps and leveraged the model to edit patient-specific label maps under various hyperparameters. By examining the the morphological and topological attributes of the label maps post-editing, we find that diffusion model-based editing techniques can generate informative morphological variants of individual digital twins. Perturbational editing can produce scale-specific variations of digital twins, which can isolate the sensitivity of device deployment to both small and large-scale variations. In contrast, localized editing can produce region-specific variations of digital twins, which can elucidate the localized effect of anatomic features on device deployment. However, we find that the generative editing process can introduce topological violations within the synthetic label maps, reducing their compatibility with numerical simulation.  Moreover, we demonstrate that diffusion models can exhibit a bias towards generating the more common anatomic features within the dataset, a bias that extends to diffusion model-based editing techniques. We nevertheless demonstrate that such anatomic bias can be leveraged to augment virtual cohorts with digital siblings for \textit{in silico} trials to improve cohort balance and diversity. Specifically, we found that perturbational editing can fill the sparsely populated regions within the anatomic distribution, potentially improving device assessment within realistic anatomies, while localized editing can expand the space of plausible anatomies that can be probed with virtual interventions, enabling the assessment of possible failure modes. 

While promising, the use of such editing techniques to augment \textit{in silico} trials should be employed with caution. Edited anatomies with low morphological plausibility can induce inaccuracies in the assessment of device safety, or fail to capture possible failure modes due to anatomic bias. Furthermore, generative editing with diffusion models can produce anatomies with topologically incorrect features, such as connected atria or several left ventricle components, which would induce non-physiological phenomena within numerical simulations of cardiovascular physics. The use of 3D convolutions within the U-net architecture further sets an upper limit on the resolution of anatomies that can be generated, limiting its use in cardiovascular contexts concerning large anatomies with small features that are critical to the fidelity of numerical simulations, such as the branching aortic vessel tree. 

Furthermore, while our experimental framework can derive novel insights regarding the morphological and topological behaviour of generative editing for virtual interventions, it exhibits a number of limitations. First, it does not quantitatively analyze morphology on multiple scales, instead measuring global level metrics such as volumes and axis lengths. Second, the influence of the diffusion model architecture on generative editing was not explored, where it is possible that the number convolutional and attention layers can influence whether a generative model learns a spatially entangled representation of anatomy. Lastly, the validity of visualizing spatial heatmaps depends on spatial correspondence between anatomic features, and would not apply to anatomies that have a variable topology such as organs with multi-component inclusions. All of these limitations present exciting directions for future work on evaluation metrics and experimental frameworks for the generative editing of digital twins.

\bibliographystyle{abbrv}
\bibliography{refs}

\appendix
\section{Appendix: Latent Diffusion Model Architecture and Training}
\label{appdx:diffusion_model}
We trained the variational autoencoder with an MSE reconstruction loss and a KL divergence loss with a relative weight of 1e-6. We modified the architecture from \cite{rombach2022ldm} to ensure compatibility with 3D voxel grids and adjusted the number of channels to [64,128,192]. We augmented our data with random scaling (0.5-1.5), rotations (0-180 degrees), and translations (0-20 voxels) in each direction. We modified the original architecture of the denoising diffusion model specified in \cite{rombach2022ldm} to ensure compatibility with 3D voxel grids and adjusted the model channels to [64,128,192]. We used the Adam optimizer \cite{kingma2014adam} for the VAE and diffusion model, using learning rates of 1e-4 and 2.5e-5 respectively. 

\begin{figure*}[t!!!]
    \centering
    \begin{minipage}{0.49\linewidth}
        \includegraphics[width=\linewidth]{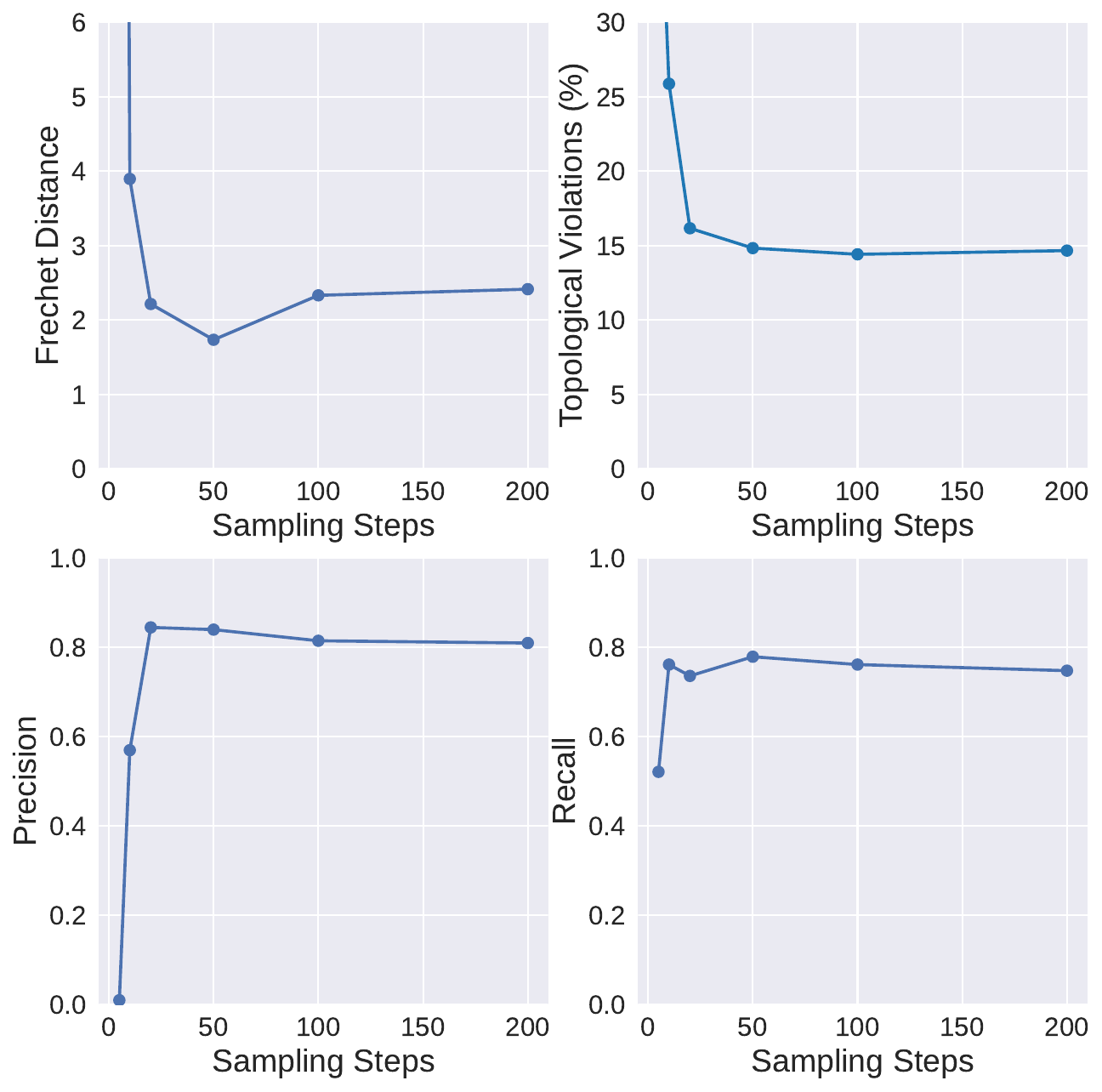}
        \caption{Lineplots demonstrating the relationship between sampling steps and virtual cohort quality.}
        \label{uncond_metrics_ksweep}
    \end{minipage}
    \hfill
    \begin{minipage}{0.49\linewidth}
        \includegraphics[width=\linewidth]{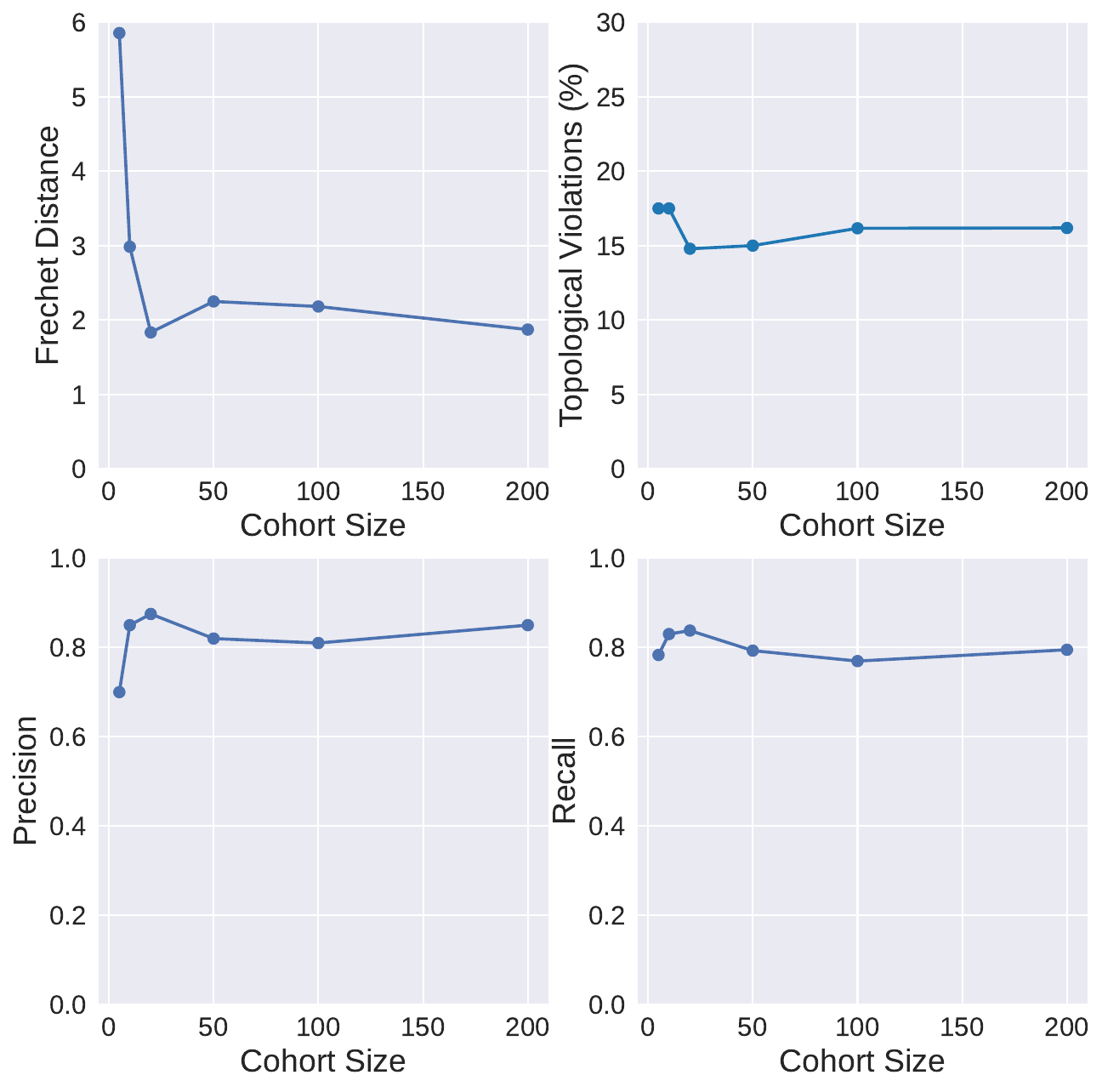}
        \caption{Lineplots demonstrating the relationship between cohort size and virtual cohort quality.}
        \label{uncond_metrics_nsweep}
    \end{minipage}
\end{figure*}

\section{Appendix: Topological Metrics}
\label{appdx: topo_metrics}
We utilize three types of metrics to assess topological violations. The first five metrics checks for the correct number of connected components for the Myo, LV, RV, LA, and RA channels. The next five metrics assesses the required adjacency relations between the following tissues: LV \& Ao, LV \& Myo, LV \& LA, RV \& Myo, RV \& RA. The final two metrics examine the absence of adjacency relations between the LV \& RV as well as the LA \& RA. Multi-component topological violations were found by determining the presence of critical voxels as described in Gupta et al. \cite{gupta2022learningtopo}.

\section{Appendix: Sensitivity Analysis for Evaluation Metrics}\label{appdx: Sens}
To determine the most efficient number of sampling steps and size of the virtual cohort, we generate six virtual cohorts with different numbers of sampling steps [5,10,20,50,100,200] and cohort sizes [5,10,20,50,100,200]. The default values of the sampling steps and cohort size was 20 and 100 respectively. We measure the Fréchet distance, precision, recall, and the percentage of topological violations for each cohort as compared to the real dataset. The results are displayed in Figures \ref{uncond_metrics_ksweep} and\ref{uncond_metrics_nsweep}. We find that measured metrics do not improve after 20 steps and a cohort size of 60, which are set as lower limits for subsequent experiments.

\end{document}